%% file: acl_latex.tex
\pdfoutput=1
\documentclass[11pt]{article}

\usepackage[]{acl}

\input{math_commands.tex}

\usepackage{times}
\usepackage{latexsym}
\usepackage{graphicx}
\usepackage{multirow}
\usepackage{color}
\usepackage{booktabs}
\usepackage{amsmath}
\usepackage{subcaption}
\usepackage[T1]{fontenc}
\usepackage[utf8]{inputenc}

\usepackage{microtype}
\usepackage{pifont}

\title{Improving Input-label Mapping with Demonstration Replay for \\ In-context Learning}

\author{ \textbf{Zhuocheng Gong\textsuperscript{1}\footnotemark[1], Jiahao Liu\textsuperscript{2}, Qifan Wang\textsuperscript{3} }\\
\textbf{Jingang Wang\textsuperscript{2}, Xunliang Cai\textsuperscript{2}, Dongyan Zhao\textsuperscript{1,4,5}\footnotemark[2], Rui Yan\textsuperscript{6}}\footnotemark[2] \\
\textsuperscript{1}Wangxuan Institute of Computer Technology, Peking University \\
\textsuperscript{2}Meituan; \textsuperscript{3}Meta AI; \textsuperscript{4}National Key Laboratory of General Artificial Intelligence \\
\textsuperscript{5}Beijing Institute for General Artificial Intelligence \\
\textsuperscript{6}Gaoling School of Artificial Intelligence, Renmin University of China \\
\texttt{\{gzhch,zhaody\}@pku.edu.cn}, \texttt{ruiyan@ruc.edu.cn}, \texttt{wqfcr@fb.com} \\
\texttt{\{liujiahao12,wangjingang02,caixunliang\}@meituan.com}}

\begin{document}
\maketitle
\begin{abstract}
% In-context learning (ICL), which solves tasks by appending a few input-label demonstrations to the input to help the model better understand the task rather than adjusting the model parameters, has become an emerging ability of large autoregressive language models. 
% The success of ICL is largely attributed to the powerful language modeling ability of LLM, which enables it to learn input-label mapping from in-context demonstrations. However, we identify a limitation in autoregressive language models that affects the performance of ICL. That is, the causal language modeling only allows for attention to previous tokens. 
In-context learning (ICL) is an emerging capability of large autoregressive language models where a few input-label demonstrations are appended to the input to enhance the model's understanding of downstream NLP tasks, without directly adjusting the model parameters. The effectiveness of ICL can be attributed to the strong language modeling capabilities of large language models (LLMs), which enable them to learn the mapping between input and labels based on in-context demonstrations. Despite achieving promising results, the causal nature of language modeling in ICL restricts the attention to be backward only, i.e., a token only attends to its previous tokens, failing to capture the full input-label information and limiting the model's performance.
% This restriction makes sense for modeling regular natural language sentences but becomes a burden when modeling input-label mapping between in-context demonstrations.
%, which means that when performing ICL, only later demonstrations can attend to earlier demonstrations, while earlier demonstrations cannot attend to later demonstrations. This results in only half of the information being utilized, which is not ideal for modeling input-label mapping between in-context examples.
In this paper, we propose a novel ICL method called Repeated Demonstration with Sliding Causal Attention, (\textsc{RdSca}). Specifically, we duplicate later demonstrations and concatenate them to the front, allowing the model to `observe' the later information even under the causal restriction. Besides, we introduce sliding causal attention, which customizes causal attention to avoid information leakage.
Experimental results show that our method significantly improves the input-label mapping in ICL demonstrations. We also conduct an in-depth analysis of how to customize the causal attention without training, which has been an unexplored area in previous research.% We believe that 

\end{abstract}

\renewcommand{\thefootnote}{\fnsymbol{footnote}}
\footnotetext[1]{Work done during an internship at Meituan.}
\footnotetext[2]{Corresponding authors: Dongyan Zhao (zhaody@pku.edu.cn) and Rui Yan (ruiyan@ruc.edu.cn).}

\section{Introduction}
Large language models (LLMs) have become the backbone of various natural language processing tasks in different fields. One of the most remarkable abilities of LLMs is in-context learning (ICL)~\citep{brown2020language}. By providing a few demonstrations and instructions into the input, along with the input queries, LLMs can perform well in new tasks without requiring fine-tuning. The secret of ICL is to formulate the input as the natural language generation task, then the LLM can be activated to prompt knowledge learned in the pre-training stage. 

Despite the promising results demonstrated by existing ICL models~\citep{DBLP:journals/corr/abs-2212-04037,DBLP:conf/nips/Wei0SBIXCLZ22}, their causal nature in language modeling restricts each token's attention solely to its preceding tokens. As a result, these models fail to capture the complete input-label information, thereby limiting their overall performance. Specifically, the pre-training objective of the current autoregressive LLMs focuses on predicting future tokens based on past ones~\cite{radford2018improving}, implemented with causal attention. While this approach works well for modeling regular sequences, it becomes less effective when applied to ICL tasks. The limitation of causal attention restricts ICL demonstrations to having only left context, which hampers the model's ability to fully comprehend and exploit the input-label relationship.

Unlike tokens in a sentence that possess sequential dependencies, there is no inherent sequential relationship between the demonstrations in the ICL input. Therefore, it is desirable for these demonstrations to interact with one another comprehensively, rather than relying solely on later demonstrations attending to earlier ones, while the reverse is not possible. Intuitively, if we can enable each demonstration to attend to all the others, we can potentially obtain a more sophisticated context for the ICL query.
However, achieving this on an LLM pre-trained with the objective of causal language modeling is not straightforward. Simply removing the causal restriction and allowing the model to have access to the right context during inference is not feasible, as it would result in a significant disparity between training and inference conditions.

% Unlike tokens in a sentence that have sequential dependencies, there is no sequential relationship between the demonstrations in the ICL input. So, we would like these demonstrations to interact fully with each other, rather than having only the later demonstrations attend to the information of the earlier demonstrations, while the earlier ones cannot attend to the later ones. Intuitively, if there is a way to let each demonstration attend to all the others, we can hopefully get a more sophisticated context for the ICL query.
% However, there is no trivial way to achieve this on an LLM pre-trained with the objective of causal language modeling. Simply removing the causal restriction and allowing the model to have the right context during inference is not feasible since it leads to a huge gap between training and inference.

%when implanting multiple demonstrations into the input, each demonstration can only attend to previous demonstrations, thus only half of the information is utilized. However, we believe such restriction is unnecessary and removable. If there is a way to let each demonstration attend to all the others, we can hopefully get a more sophisticated context. 

In this work, we focus on capturing full input-label mapping information from demonstrations. To achieve this target, we propose two techniques. The first is \textbf{Repeated Demonstration}, where we replicate later demonstrations and concatenate them to the front. This allows the model to `observe' the later information even under the causal restriction. 
For example, if we consider four demonstrations represented by $d_1d_2d_3d_4$, the input sequence after replication becomes $d_2'd_3'd_4'd_1d_2d_3d_4$.
However, simply duplicating demonstrations brings about a new problem: we do not want to attend a demonstration twice, as this may cause the model to take shortcuts by learning to repeat the answer of its first encounter, rather than learning the input-label mapping. To address this, we propose the second technique, the \textbf{Sliding Causal Attention}, which customizes the original causal attention by restricting the attention window so that each demonstration can only attend to all other demonstrations once. In the case of four demonstrations, attention windows are $d_2'd_3'd_4'd_1$, $d_3'd_4'd_1d_2$, $d_4'd_1d_2d_3$, and $d_1d_2d_3d_4$, respectively. Through experiments, we demonstrate that our proposed method (Repeated Demonstrations with Sliding Causal Attention, \textsc{RdSca}) significantly enhances the ability to learn input-label mapping from ICL demonstrations. 
 
Our proposed sliding casual attention is the first attempt that customizes the causal attention in the inference stage without further training. We investigate a number of different designs for customizing causal attention and reach some principles for the success of ICL. For example, we find that the first \texttt{<SOS>} token plays an essential role. It should always be available to attend to no matter where the attention window slides. Besides, the size of the attention window determines the richness of the semantic context, thus it affects the performance greatly.
%As can be seen, all demonstrations appear exactly once in each attention scope.
%Previous research has shown that the ability of ICL comes from two sources: the semantic prior of LLM and the input-label mapping of the demonstrations. 
%To validate this, we remap the natural language label words in the demonstrations to irrelevant symbols so that models cannot rely on relevant natural language label words to induce the semantic prior, which we call \textit{symbolic ICL}. Under this setting, we experiment on a series of models with different scales on various tasks and witness significant performance improvements, demonstrating that RDSCA better captures the input-label mapping information in the demonstrations.
Our contributions are summarized as:
\begin{itemize}
    \item %To the best of our knowledge, we are the first to question the rationality of causal language modeling in the ICL scenario and explore ways to enhance the interaction between demonstrations. 
    To the best of our knowledge, we are the first to identify the limitation of causal language modeling in the ICL and to introduce a novel approach for enabling effective interactions between demonstrations.
    \item We validate the feasibility of customizing causal attention during the inference stage without further training and conduct further analysis on causal attention customization. We believe this idea has great potential and sheds new light on optimizing ICL and other large model inference scenarios.
    \item %We conduct experiments on  various text classification datasets. Experimental results show that our proposed method significantly improves the input-label mapping in ICL demonstrations.
    We conduct experiments on several text classification datasets to evaluate the effectiveness of our proposed method. The experimental results clearly demonstrate that our approach significantly enhances the input-label mapping in ICL demonstrations.
\end{itemize}

\begin{figure*}[t]
  \centering
  \includegraphics[width=1\textwidth]{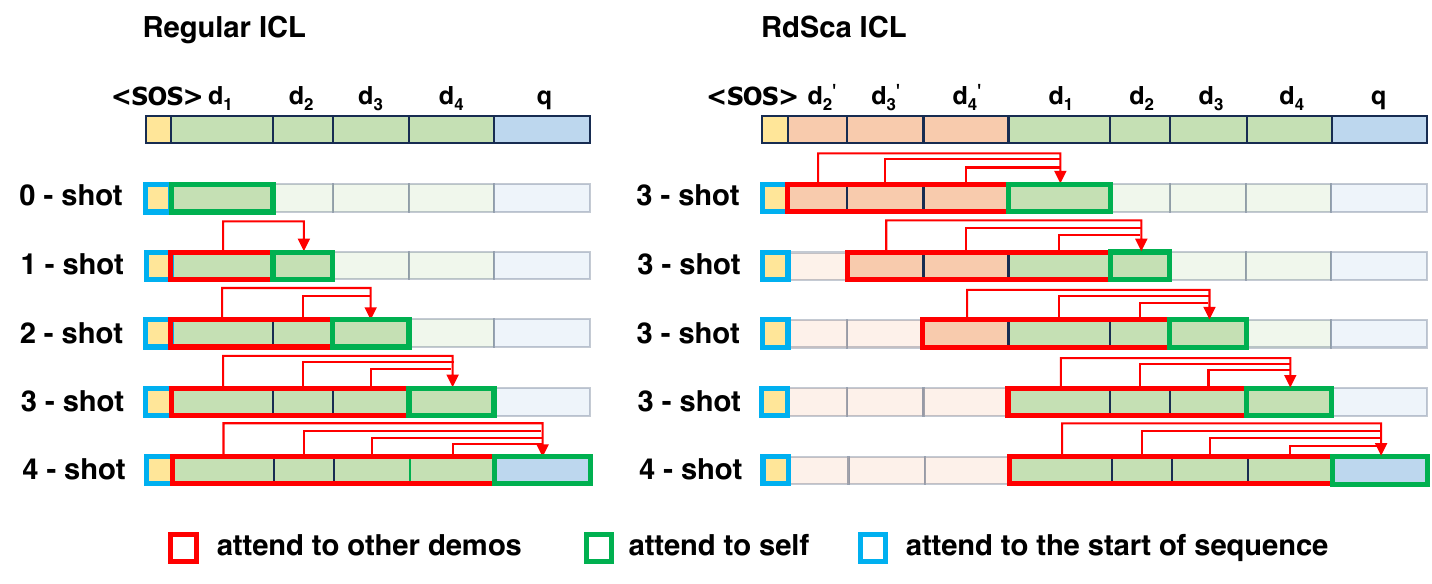}
  \caption{An illustration for regular ICL and our \textsc{RdSca} ICL. Regular ICL uses vanilla causal attention, causing the uneven interaction between demonstrations. In \textsc{RdSca}, each demonstration can attend to all the others thanks to demonstration repetition and sliding causal attention.}
  \label{fig:main}
  \vspace{-3mm}
\end{figure*}

\section{Backgrounds}
\subsection{Causal Language Modeling}
Most of the current decoder-only LLMs employ causal language modeling as the pre-training objective~\citep{radford2018improving}, which aims to predict future tokens based on the past.  
%Causal language modeling aims to predict the token following a sequence of tokens.
\begin{equation}
    \operatorname*{arg\,min}_{\theta}\sum\limits_{i}P_{\theta}(u_i|u_0,u_1,...,u_{i-1})
\end{equation}
In causal language modeling, the model only attends to tokens that occur before (the left context), resulting in a unidirectional attention scheme known as \textbf{causal attention}.
This approach enables the model to process each token in the input sequence in order, without accessing any information from the future (the right context).
\paragraph{Causal Attention Mask} Practically, a causal attention mask is used to implement causal attention, which guarantees unidirectionality by masking all right-to-left attention connections and only allowing right-to-left connections.
Formally, the attention mask is a binary-valued matrix $M\in\{0,1\}^{n\times n}$, where $n$ is the total sequence length. The element $m_{ij}$ in $M$ indicates whether the $j$-th token in the sequence can attend to the $i$-th token, with a value of $1$ for yes and $0$ for no. Therefore, the causal attention mask matrix is a lower triangular matrix where $m_{ij}=0,\forall i<j$.

\subsection{In-context Learning}
Now we formally introduce the definition and basic notations of in-context learning. We focus on classification tasks and causal language models. Given $K$ input-label pairs $\{x_i, y_i\}_{i=1}^K$ and a query $x_q$, the objective is to predict the label of the query $y_q$ by predicting the next token. Formally,
\begin{equation}
    \operatorname*{arg\,min}_{y_q\in \mathcal{C}}P(y_q|(x_1,y_1),(x_2,y_2),...,(x_K,y_K),x_q)
\end{equation}
where $\mathcal{C}$ is the label set.

%\paragraph{Prompts and Templates} 
To perform ICL successfully, we transform the classification task into a natural language generation task by adding templates to the query and demonstrations. Additionally, discrete labels are mapped to label words, such as "positive" and "negative" for sentiment classification.
We denote the demonstration with templates and label words as $d_i=T(x_i,y_i)$. %Concatenating demonstrations and the query $q=T(x_q,_)$ forms the whole input to the model $d_1d_2...d_kq$
The entire input to the model is formed by concatenating the demonstrations and the query $q=T(x_q,\_)$, resulting in $d_1d_2...d_kq$.
%Given demonstrations and the query, the first thing that needs to be done is to add prompts to the input, which converts the classification task to the natural language generation task. We also need to map discrete labels to label words (e.g., the task of sentiment classification uses "positive" and "negative" as label words). 
\paragraph{Semantically-unrelated label ICL (SUL-ICL)} In regular ICL, natural language words that are closely related to the task objective are used as label words. This allows the model to utilize semantic prior knowledge. For example, when conducting sentiment classification, label words such as "positive" and "negative" are highly semantically related to the sentiment labels of the input. During pre-training, the model is exposed to similar patterns and can learn to associate these label words with the corresponding sentiment labels.
%, so that LLMs can successfully utilize semantic prior knowledge to perform in-context learning successfully. 
In this paper, we eliminate the contribution of semantic priors and perform semantically-unrelated label ICL~\citep{wei2023larger}. In this setting, natural language labels are replaced with semantically-unrelated labels. This approach forces the model to rely solely on input-label mappings to perform ICL.

%\textbf{Right}: Causal attention of regular ICL. \textbf{Middle}: Customized causal attention of \textsc{RdSca}. Practically, causal attention is implemented by applying masks to the attention matrix, which we denote with grey color. The red block represents the attention window. Each attention window contains a fixed amount of demonstrations. \textbf{Left}: A comparison between regular ICL and our model.

% As discussed in~\citet{wei2023larger}, two factors contribute to the success of ICL: (a) semantic prior knowledge, which involves using natural language label words such as "positive" and "negative" and performing ICL using prior knowledge, and (b) input-label mappings from the presented demonstrations.
% We find that the current paradigms of casual language modeling may not fully exploit the potential of input-label mappings in ICL.

\section{Method}
\subsection{Defect of Traditional ICL}
% When performing few-shot ICL, we concatenate the labeled demonstrations with the query and input them into the LLM. The labeled demonstrations provide in-context input-label information, which is why few-shot ICL usually outperforms zero-shot ICL.
During few-shot ICL, we combine the labeled demonstrations with the query by concatenating them and feeding them into the LLM. The labeled demonstrations provide valuable in-context input-label information, which is included in the few-shot ICL input. As a result, few-shot ICL consistently achieves better performance compared to zero-shot ICL.

However, few-shot ICL under the current scheme of causal language modeling has a defect. Due to the restriction of causal attention, each demonstration only has access to half of the full context (\emph{i.e.}, the left context). As a result, it cannot `observe' demonstrations that follow it. For example, in 4-shot ICL, if we consider the third demonstration as the last query (which makes sense because tokens that come after it have no influence on it), then the first two demonstrations serve as its context while the fourth demonstration is ignored. In this sense, we can regard predicting the label of the first demonstration as a zero-shot ICL task, the second as a one-shot ICL, the third as a two-shot ICL, and so on, as shown in Figure~\ref{fig:main}.

While restricting the model from accessing the right context makes sense when modeling regular natural language sequences to prevent information leakage, it is unnecessary for few-shot ICL since there is no dependency between demonstrations. The causal restriction limits earlier demonstrations from accessing later information, resulting in only half of the available information being utilized.

%To address this issue, we need to explore alternative modeling approaches that can effectively capture the information from all demonstrations without being restricted by the causal attention mechanism.

\subsection{Repeated Demonstrations}
Enabling each demonstration to `observe' later demonstrations is not a trivial task. Simply replacing causal attention with full attention to allow the model to receive both left and right context is not feasible, as the model is pre-trained with causal language modeling. Switching to full attention during inference would result in a significant loss in performance, as we have verified through experiments (Section~\ref{sec:full}). Therefore, we propose the \textbf{Repeated Demonstration} method to establish sufficient interactions among demonstrations while still adhering to the premise of causal attention. The method is based on a simple idea: duplicating all the other demonstrations except the first one and concatenating them to the front of the sequence. By doing so, we expand the input sequence $d_1d_2...d_Kq$ to $d_2'...d_K'd_1d_2...d_Kq$ where $d_i'$ is the duplication of $d_i$. This operation allows each demonstration to attend to all the others by adding the later demonstrations to its left context. As a result, the model can make full use of demonstrations while still obeying the causal attention restriction.

\subsection{Sliding Causal Attention}
The Repeated Demonstration method alone is not sufficient to achieve our target, as it introduces a new problem of duplicated information in the context. Under causal attention, a token can attend to all tokens that precede it. When combined with repeated demonstrations, this can result in the same information being duplicated in the context. For example, when performing 4-shot ICL with repeated demonstrations, the input is $d_2'd_3'd_4'd_1d_2d_3d_4q$. The context of $d_3$ is $d_2'd_3'd_4'd_1d_2d_3$ where some demonstrations appear twice. 
Repetitive information can cause the model to learn a shortcut to predict the label by repeating the label of the same demonstration that appeared for the first time, rather than learning the input-label mapping as expected. We will provide a detailed explanation of this phenomenon in the experimental section (Section~\ref{sec:custom}).

% Therefore, our goal is clear: for each demonstration, we want other demonstrations to appear only once in the context.
To tackle this issue, we propose \textbf{Sliding Causal Attention}, which utilizes a sliding attention window to limit the range of tokens that each demonstration can attend to. This approach effectively prevents the occurrence of repetitive information and ensures that each demonstration has a non-repetitive context. Specifically, the attention window is defined as follows:
\begin{equation}
    window(x) = \begin{cases}
                d_{i+1}'...d_{K}'d_{1}...d_{i} & x=d_i, \forall i \le K\\
                d_{1}d_2...d_{K}q & x=q
           \end{cases}
\end{equation}

Then we explain how sliding causal attention works. We use the window size $W$ to represent the number of demonstrations contained within it. In our main setting, we use $W=K$, indicating each attention window contains all the demonstrations. As the window slides, a new demonstration enters while an old one exits. This ensures that there are always consistent $K$ different demonstrations within the context.

Additionally, we find that the \texttt{<SOS>} token, which represents the first token of the sequence, is crucial to the model's performance.
Therefore, we add \texttt{<SOS>} to every attention window to ensure that each token can attend to \texttt{<SOS>}. We will provide further explanations on this in the experimental section~\ref{sec:sos}.

\begin{figure*}[ht]
  \centering
  \includegraphics[width=1.\linewidth]{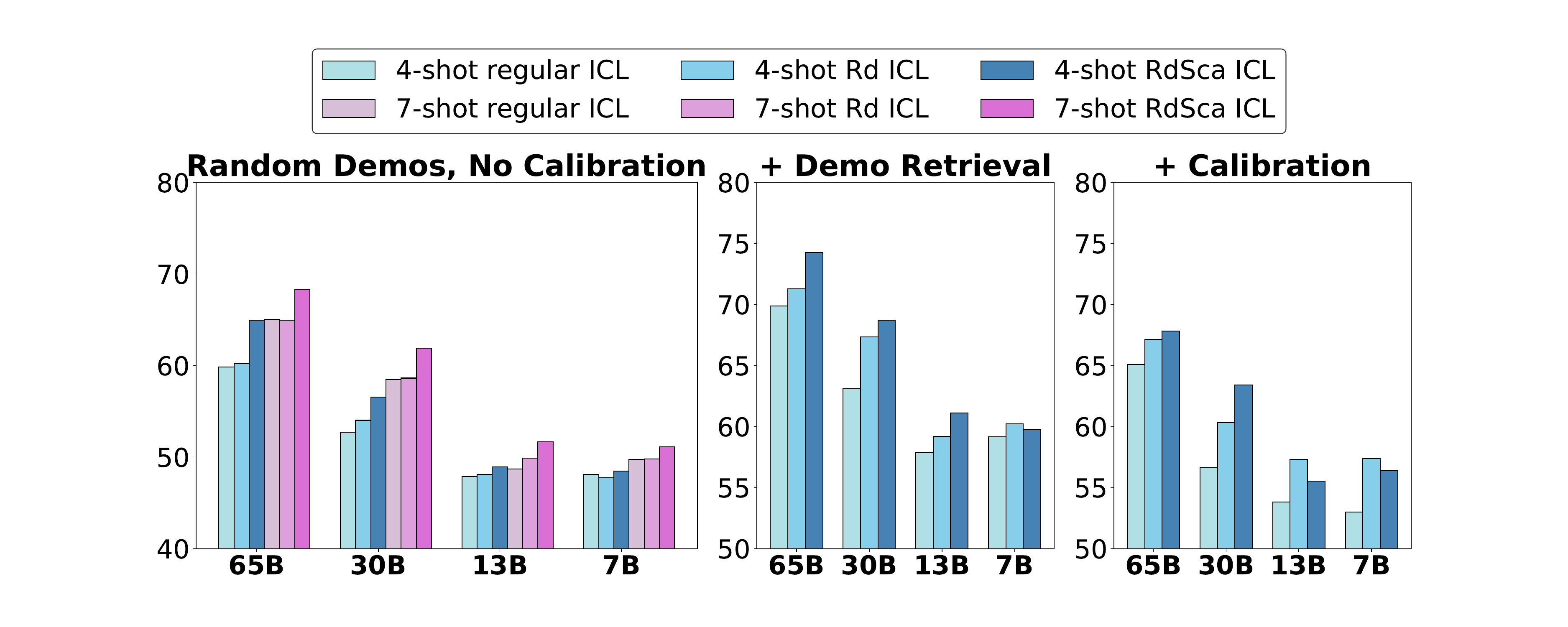}
  \vspace{-8mm}
  \caption{Averaging performance of \textsc{RdSca} on LLAMA of different scales. \textbf{Left}: ICL with random-selected demonstrations. We report the results of both 4-shot and 7-shot ICL for this setting. \textbf{Middle}: We combine our method with the demonstration retrieval technique. We employ \textsc{KATE}~\citep{DBLP:conf/acl-deelio/LiuSZDCC22} to retrieve target demonstrations. \textbf{Right}: we combine with \textsc{Prototypical Calibration}~\cite{DBLP:journals/corr/abs-2205-10183} to further improve performance. Full results can be found in Appendix~\ref{appendix:full}}
  \label{fig:res}
  \vspace{-5mm}
\end{figure*}

\section{Experiments}
\subsection{Setup}
\paragraph{Models} We conducted experiments on decoder-only causal language models. We utilized the LLAMA~\citep{touvron2023llama} model family with varying scales, including 7B, 13B, 30B, and 65B.
\paragraph{Tasks} We evaluate on classification tasks, including \textbf{SST-2}~\citep{socher2013recursive}, \textbf{CB}~\citep{de2019commitmentbank}, \textbf{RTE}~\citep{dagan2006pascal,wang2019superglue}, \textbf{AGNews}~\citep{zhang2015character}, \textbf{QQP}~\citep{DBLP:conf/iclr/WangSMHLB19}, and \textbf{QNLI}~\citep{DBLP:conf/iclr/WangSMHLB19}. 
If the dataset includes a validation split, we evaluate the model's performance on the validation set. Otherwise, we evaluate on the test set. Datasets are obtained from Huggingface \texttt{Datasets} library\footnote{https://huggingface.co/docs/datasets/index}.
\paragraph{Other Details}
For all experiments, we use $K$ = 4 demonstrations by default. Demonstrations are uniformly sampled from the training data.
We utilize prompt templates from \texttt{PromptSource}\footnote{https://github.com/bigscience-workshop/promptsource}~\citep{bach2022promptsource}.
For each dataset, we use four different templates and select a set of $K$ training examples using 4 different random seeds. Therefore, the reported results are the average of 16 different runs.
We would like to emphasize that we run all methods using the same random seeds, ensuring that the demonstrations used by each method are identical. This was done to eliminate any potential bias caused by differences in the demonstrations seen by each method and ensure a fair comparison. Due to limited computing resources, we evaluate a random sample of 200 examples for tasks with a large validation set, instead of the entire dataset. We have found that this sample size is sufficient to obtain stable and reliable results.

\subsection{Main Results}
Figure~\ref{fig:res} presents a comparison between our method and regular ICL. We observe a clear correlation between the performance of semantically unrelated label ICL and the model scale: increasing the model scale improves performance for all methods. In the case of LLAMA-7B, the performance is poor, and the performance gap between different methods is small, indicating that it is challenging for small models to learn the input-label mapping from demonstrations while the semantic meaning of labels is removed. As the model size scales up to 30B and 65B, the models begin to exhibit strong ICL capabilities, and the performance of different methods varies. This phenomenon is consistent with the findings of~\citet{wei2023larger}, which suggest that larger language models learn the input-label mapping differently because the input-label mapping learning ability only emerges when the model is large enough.
In comparison with ICL with random demonstrations, we find that retrieving demonstrations that are more relevant to the query can significantly improve the performance on all scales. Furthermore, calibration has a positive influence on the ICL results, showing that our method can be combined with other techniques to further enhance the overall performance.

\paragraph{\textsc{RdSca} improves input-label mapping}
%加点数据
We observe a significant performance boost over regular ICL in both 4-shot and 7-shot settings on LLAMA-30B and LLAMA-13B. Our method shows an average improvement of 8.4\% on LLAMA-30B and 10.5\% on LLAMA-65B compared to regular ICL, indicating a significant advantage in learning the input-label mapping from demonstrations.
As mentioned in previous sections, we believe that in the scheme of causal language modeling, regular ICL only utilizes half of the context (left context) for the demonstrations. In contrast, our method makes use of all the available information, which enhances the representations of demonstrations and provides richer semantics for predicting the query's target.% Our experiments have demonstrated the effectiveness of our method.

%As mentioned in previous sections, we believe that in the scheme of causal language modeling, regular ICL only uses half of the context (left context) for the demonstrations. Our method makes use of all the available information, which enhances the representations of demonstrations and thus provides richer semantics for predicting the query's target. Our experiments demonstrate the effectiveness of our method.

\subsection{Results on MMLU}
\input{tables/mmlu}
We also validate LLAMA-65B on MMLU~\cite{DBLP:conf/iclr/HendrycksBBZMSS21}, which is a more advanced benchmark for evaluating the abilities of LLMs. The results are shown in Table~\ref{mmlu}.

\begin{figure}[t]
  \centering
  \includegraphics[width=0.4\textwidth]{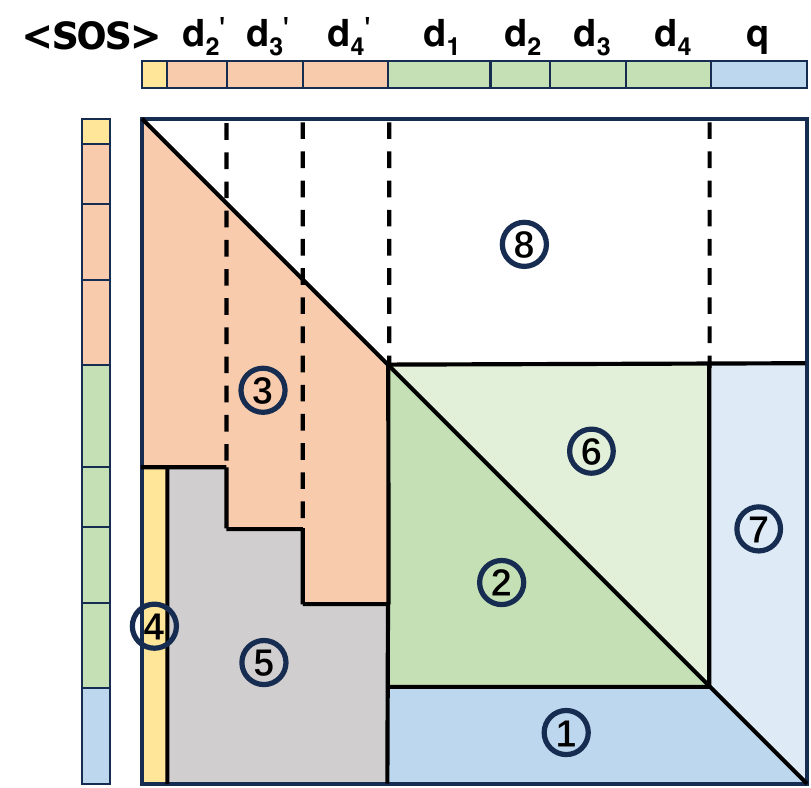}
  \caption{Illustration of the attention mask, which we divide into multiple regions and label with numbers. The specific meanings are as follows: \ding{172} query-to-demonstrations attention; \ding{173} inter-demonstrations right-to-left attention; \ding{174} inter-demonstrations(repeat) right-to-left attention; \ding{175} attention to the start-of-sentence token; \ding{176} attention to redundant context; \ding{177} inter-demonstrations left-to-right attention;\ding{178} demonstrations-to-query attention;\ding{179} other left-to-right attention.}
  \label{fig:attn}
  \vspace{-5mm}
\end{figure}
\input{tables/attention}

\subsection{Looking Deeper into Causal Attention Customization}
In this section, we conduct a further investigation to look deeper into the effectiveness of sliding causal attention. \textit{How and why does sliding causal attention work and are there alternatives when customizing causal attention?} To answer these questions, we implement the following ablations:

\begin{itemize}
   % \item \textsc{Regular ICL}. This is the regular ICL setting.
    \item \textsc{Full Attn.} This uses full attention instead of causal attention so the model can attend to both the left and right context.
    \item \textsc{Rd}. This is the ablation of \textsc{RdSca} that uses original causal attention without sliding causal attention.
    \item \textsc{RdSca} w/o \texttt{<SOS>}. In this ablation of \textsc{RdSca}, when the attention window moves forward, latter tokens cannot attend to the beginning of the sentence, \emph{i.e.}, the \texttt{<SOS>} token.
\end{itemize}
Figure~\ref{fig:attn} shows how we divide the whole attention matrix into various regions. The above-mentioned methods have unique access to these regions. Details and experimental results are shown in Table~\ref{tab:attn}.

\paragraph{Can we break the causal restriction?}
\label{sec:full}
First, we are concerned about the limitations of causal attention. Intuitively, if we can remove the causal restriction during inference by using non-causal attention and providing the model with the full context, we can fully utilize the demonstration information. However, our experimental results indicate that this simple idea is not feasible. The performance of \textsc{Full Attn.} is no better than random guessing, but why?
There is a simple explanation: the gap between causal attention training and non-causal attention inference is too huge. Since the model is trained using causal language modeling, it has no inductive bias toward data with the correct context. Therefore, using full attention during inference is out-of-distribution for a causal model.

\paragraph{Sliding attention window matters.}
\label{sec:custom}
Based on the results, it is evident that \textsc{Rd} has little improvement over regular ICL. Although \textsc{Rd} provides richer interaction between demonstrations than regular ICL, the lack of attention restrictions causes the same demonstration to attend to its first appearance. Consequently, the model takes a shortcut and repeats the answer of the first appearance instead of learning the input-label mapping to predict the result. Therefore, the improvement brought by simply repeating the demonstrations is limited. However, the situation changes when we add the sliding causal window, which blocks repeated information and forces the demonstrations to attend only to other unseen demonstrations, ensuring that there is no information leakage. According to the evaluation results, this approach successfully enables the model to better capture the input-label mapping information, as expected.

This ablation study further indicates another insight: the default causal attention may not be always optimal in ICL. Previous research aimed at improving ICL has mainly focused on template construction and demonstration selection, with little attention paid to the causal attention masks. Our study shows that customizing attention masks can be a new technique for enhancing ICL, which is worthy of further research. 
%By modifying the attention masks, we can control the information flow and improve the model's ability to capture the input-label mapping information. Therefore, future research should explore the potential of customizing attention masks to improve ICL performance.

\paragraph{Attending to the \texttt{<SOS>} token is essential.}
\label{sec:sos}
Next, we examine the role of the first token \texttt{<SOS>} in causal attention customization. If not treating the \texttt{<SOS>} token separately, as the attention window moves, the first token slides out of the scope of the attention window. This means that the latter demonstrations and the query cannot attend to \texttt{<SOS>}. Our experiments show that in this setting the ICL ability is severely affected (\ding{172}+\ding{173}+\ding{174} in Table~\ref{tab:attn}). However, when we manually add the \texttt{<SOS>} token to the attention window (as in the case of \textsc{RdSca}), we observe a significant improvement in performance. This comparison demonstrates that the \texttt{<SOS>} token is crucial for the model to perform correctly.
But why does such a seemingly insignificant token have such a significant impact on performance? We believe that the significance of the \texttt{<SOS>} token lies in allowing the model to treat the tokens under the attention window as a valid sequence rather than a fragment of the whole sequence.

As the model is pre-trained on sequences that all start with \texttt{<SOS>}, it does not know how to handle sequences that do not start with \texttt{<SOS>}. Therefore, when customizing causal attention, we need to make sure that the model regards the tokens in the attention window as a valid sequence starting with \texttt{<SOS>}. Only in this way, we can activate the model's language modeling ability correctly.

\input{tables/window}

\subsection{Ablation on Attention Window Size}
In this section, we discuss the impact of window size. 
We denote the attention window size of the original \textsc{RdSca} as $W=K$, meaning that there are $K$ demonstrations in each window. In this case, each demonstration can attend to all other ones, so each demonstration can be considered as performing ($K-1$)-shot ICL.
%, and the query can attend to all k examples, so the query prediction is k-shot ICL. 
As shown in Figure~\ref{fig:window}, we employ smaller window sizes $W$ on \textsc{RdSca} and see what happens to the ICL ability.
Intuitively, reducing the size of the sliding window means reducing the context (the number of demonstrations) that the current one is able to attend to. Thus the model's ability to learn the input-label mapping from context would be affected. Especially, when $W=1$, each demonstration can only attend to itself, which is equivalent to zero-shot ICL. The results are shown in Table~\ref{tab:window}. As expected, the window size is highly correlated with ICL performance: the larger the window size, the better the performance. Surprisingly, we notice that when $W=2$, \emph{i.e.}, one-shot ICL for all demonstrations, \textsc{RdSca} is already comparable with regular ICL on both 30B and 65B LLAMA. This indicates that our ICL method makes more efficient exploitation of the demonstrations in ICL.
The last row of Table~\ref{tab:window} shows when the query not seeing all 4 demonstrations, the performance drops.

\subsection{What If Adding More Demonstrations?}
\begin{figure}[ht]
  \centering
  \includegraphics[width=1.\linewidth]{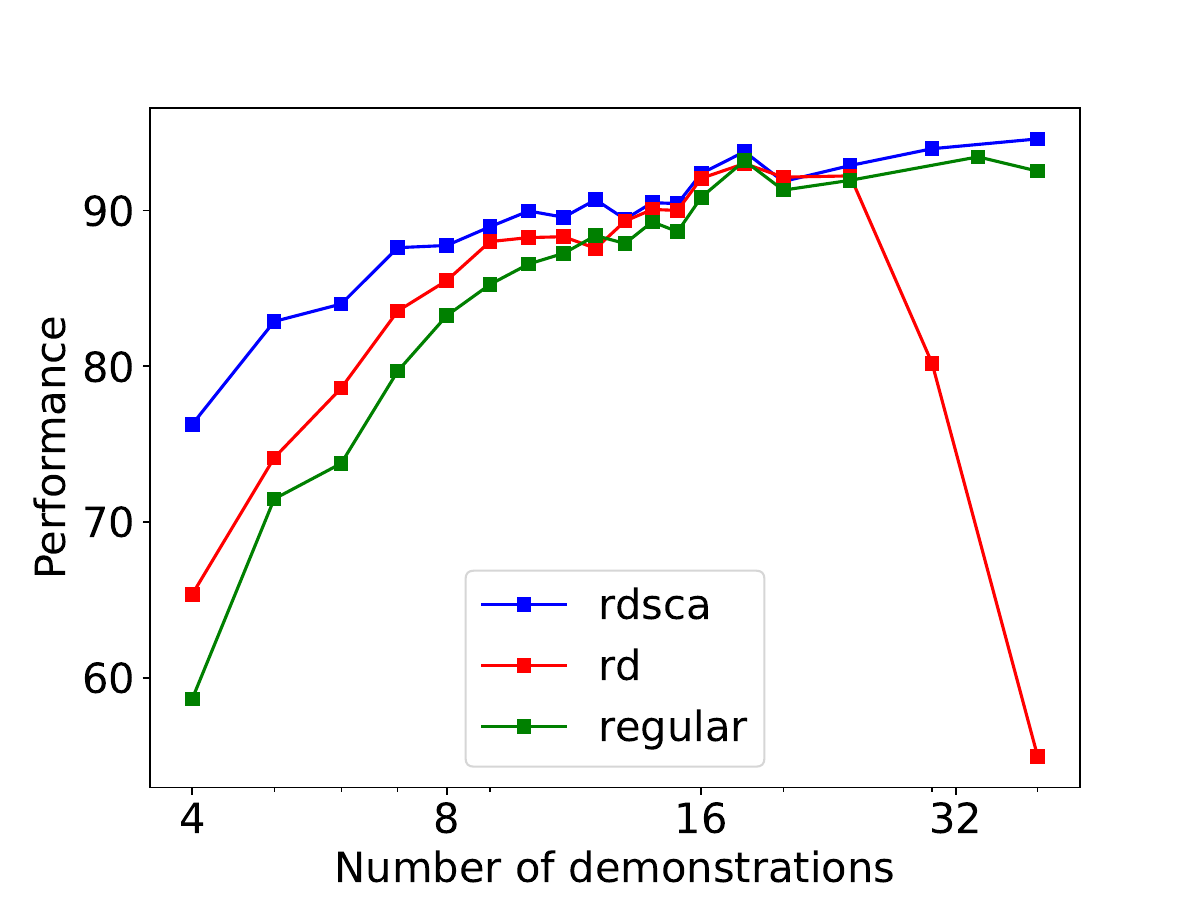}
  \caption{Increasing the number of demonstrations for LLAMA-65B on SST-2.}
  \label{fig:seqlen}
\end{figure}
%Previous works have shown that adding more demonstrations does not always contribute positively to the performance of ICL~\cite{}. 
In this section, we investigate how our method performs with more demonstrations. We start from $K=4$ demonstrations and gradually add more. As shown in Figure~\ref{fig:seqlen}, all methods receive significant performance gains when the number of demonstrations is relatively small. However, as the number of demonstrations increases, the benefits of adding more demonstrations gradually reduce. The ICL performance reaches a plateau at around $K=16$. Before reaching this plateau, \textsc{RdSca} consistently outperforms the vanilla-repeat and regular ICL, indicating that our method is robust to the increase in the number of demonstrations. However, this advantage gradually diminishes when adding more demonstrations. All methods seem to perform comparably at around $K=16$, suggesting that the ability to learn from context is close to its upper bound at this point.%, which is consistent with previous research findings~\cite{}.

In addition, we observe an interesting phenomenon in the experiment. We find that the performance of \textsc{Rd} sharply declines when there are too many demonstrations (decreasing by 12.02 points from $K=24$ to $30$, and 25.24 points from $K=30$ to $40$.), while \textsc{RdSca} does not suffer such huge losses. 
%The only difference between RD and RDSCA is whether to use a sliding causal attention window to constrain the scope of attention, which leads to such a significant performance difference. 
We explain this phenomenon from two aspects: First, as mentioned earlier, if not customizing the causal attention, the model learns shortcuts from repeated examples rather than input-label mapping, which leads to a decline in ICL performance. Second, due to the repetition of demonstrations, the context becomes too long when adding too many demonstrations, and some studies have shown that LLMs still have difficulty modeling long sequences. Our method can effectively solve these problems by customizing attention. Therefore, we believe that our exploration of customizing causal attention highlights a possible solution for tackling the long sequence modeling issue, which is worth further research in the future.

%\subsection{Ablations on Templates}
\subsection{Discussions on Efficiency Issues}
As repeating demonstrations involves expanding the sequence lengths, the model is required to process more tokens when compared with regular ICL, which may lead to computational inefficiency and extra memory consumption. However, we argue that these efficiency issues can be neglected practically.
Thanks to the auto-regressive nature of Large Language Models LLMs, the representations of previous tokens are not dependent on later tokens. Therefore, in the inference phase, the representations of demonstrations can be pre-computed and stored offline, thereby mitigating the added computational burden. 
Moreover, our adaptation of the attention mask allows the LLM to focus only on tokens within the sliding window for key-value (KV) caching, rather than retaining information on all previous tokens. This optimization reduces the memory requirements of RdSca. Consequently, the memory consumption of RdSca remains on par with that of a standard ICL setup.

\section{Related Works}
\subsection{In-context Learning}
In-context learning (ICL) is an effective approach for adapting pre-trained LLMs to downstream tasks~\cite{brown2020language}. This is achieved by adding task-specific templates and demonstrations before the test query, without updating model parameters. Recent works focus on enhancing ICL with various techniques. For example, \citet{DBLP:conf/acl-deelio/LiuSZDCC22,DBLP:conf/naacl/RubinHB22, DBLP:journals/corr/abs-2302-00083,DBLP:journals/corr/abs-2305-14128} propose that choosing demonstrations more carefully results in better ICL performance. Some studies try improving ICL with more sophisticated prompt templates, either by hand-craft or by automation~\citep{DBLP:journals/corr/abs-2212-04037}. Some propose chain-of-thoughts (CoT) to elicit the reasoning abilities of LLMs by augmenting each demonstration with a chain of reasoning steps~\citep{DBLP:conf/nips/Wei0SBIXCLZ22}. Subsequent studies have continued to build upon this approach, achieving further improvements~\citep{DBLP:journals/corr/abs-2203-11171}.
% Another line of research focuses on understanding why ICL works.....
%\paragraph{The role of demonstrations in ICL} 
Some studies explore the mechanism of ICL. \citet{DBLP:conf/emnlp/MinLHALHZ22} shows that randomly replacing labels in the demonstrations barely hurts performance so ground truth demonstrations are not required. \citep{wei2023larger} claims that the success of ICL relies on both semantic prior knowledge and input-label mappings. They find that larger models can better capture input-label mappings from demonstrations. 

\subsection{Attention Customization}
There has been a number of work done on attention customization since Transformer was proposed~\citep{DBLP:conf/nips/VaswaniSPUJGKP17}. many of them focus on modeling long sequences~\citep{DBLP:conf/iclr/KitaevKL20,DBLP:journals/corr/abs-2004-05150} or improving efficiency through sparse attention~\citep{DBLP:journals/csur/TayDBM23}. Some works have explored customizing attention to block certain dependencies. For example, \citet{DBLP:journals/corr/abs-2304-08467} prevents later tokens from attending to the prompt to compress context.
These attention customizations can be viewed as some kind of modifications to the model architecture, which requires additional training. However, our study is the first to investigate the possibility of attention customization during inference without further training.

\section{Conclusion}
In this study, we introduce \textsc{RdSca}, a novel ICL framework that enhances the learning of the input-label mapping from demonstrations. We propose that the causal attention mechanism of decoder-only LLMs restricts the model from fully exploiting the input-label mapping from demonstrations. To address this, we suggest repeating demonstrations to allow each demonstration to have full context and customizing the vanilla causal attention mechanism to prevent information leakage. Experimental results show that our method consistently improves the input-label mapping ability of ICL on LLMs of different scales. Furthermore, we delve deeper into causal attention customization and show how different attention settings affect ICL performance. Additionally, this work is the first to customize the causal attention of a pre-trained autoregressive LLM without further training, which may pave the way for further research in this direction.

\section*{Limitations}
Although \textsc{RdSca} has demonstrated outstanding performance in capturing the full input-label mapping from demonstrations, there are some limitations to our work in this section that we must acknowledge. First, our method expands the sequence length by duplicating demonstrations, which leads to increased computation costs and inference latency. This may limit its practicality in computation-constrained scenarios.
Moreover, we have not evaluated our method on more complex tasks, nor have we determined its performance with chain-of-thought scenarios. Therefore, in the future, we need to investigate how it can be extended to a broader range of task scenarios.

\section*{Acknowledgments}

This work is supported by National Key R\&D Program of China (No. 2021YFC3340304) and National Natural Science Foundation of China (NSFC Grant No. 62122089). Jingang Wang is funded by Beijing Nova Program (Grant No. 20220484098). We sincerely thank all reviewers for their valuable comments and suggestions, which are crucial for improving our work.

\bibliography{anthology,custom}
\bibliographystyle{acl_natbib}

\clearpage

\appendix
\section{Full results}

\label{appendix:full}
\input{tables/appendix_full}

\end{document}

%% file: math_commands.tex
\usepackage{amsmath,amsfonts,bm}

\def\eqref#1{equation~\ref{#1}}
\def\1{\bm{1}}

\DeclareMathAlphabet{\mathsfit}{\encodingdefault}{\sfdefault}{m}{sl}
\SetMathAlphabet{\mathsfit}{bold}{\encodingdefault}{\sfdefault}{bx}{n}

%% file: tables/mmlu.tex
\begin{table}[h]
    \centering
      \scalebox{1}{\begin{tabular}{lll}
        \toprule
         & \textsc{Regular ICL} & \textsc{RdSca}  \\
        \toprule
        humanities & 65.99 & 68.44\\
        STEM & 51.84 & 51.57\\
        social sciences & 72.29 & 73.60\\
        other & 63.34 & 64.65\\ \midrule
        average  & 62.20 & 63.66\\
        \bottomrule
      \end{tabular}}
      \caption{LLAMA-65B 4-shot ICL results on MMLU.}
      \label{mmlu}
\end{table}

%% file: tables/attention.tex
\begin{table*}[t]
\begin{center}
\setlength{\tabcolsep}{1.2mm}{
\scalebox{1}{\begin{tabular}{lllrlrl}
\toprule
Methods & Attention Regions & Is causal? & \multicolumn{2}{c}{LLAMA-65B} & \multicolumn{2}{c}{LLAMA-30B}  \\
\midrule
\textsc{Random Guess} & - & - & $43.06$& & $43.06$ \\
\midrule
\textsc{Regular ICL} & \ding{172}+\ding{173} & Yes & $59.83$ & ($+16.77$) & $52.71$ & ($+9.65$)\\
\textsc{Full Attn.} (v1) & \ding{172}+\ding{173}+\ding{177} & No  & 45.15 & ($+2.09$) & 47.84 & ($+4.78$) \\
\textsc{Full Attn.} (v2) & \ding{172}+\ding{173}+\ding{177}+\ding{178} & No & $44.32$ & ($+1.26$) & $47.36$ & ($+4.30$) \\
\midrule
\textsc{Rd} & \ding{172}+\ding{173}+\ding{174}+\ding{175}+\ding{176} & Yes & $60.18$ & ($+17.12$) & $54.02$ & ($+10.96$)\\
\textsc{RdSca} w/o \texttt{<SOS>} & \ding{172}+\ding{173}+\ding{174} & Yes & $32.20$ & ($-10.86$) & $12.96$ & ($-30.10$)\\
\midrule
\textsc{RdSca}  & \ding{172}+\ding{173}+\ding{174}+\ding{175} & Yes & $64.96$& ($+\textbf{21.90}$) & 56.55 & ($+\textbf{13.49}$)\\
% \midrule
% & \ding{172}+\ding{173}+\ding{177}+\ding{178}+\ding{179} \\
\bottomrule
\end{tabular}}}
\end{center}
\caption{Ablations on causal attention customization. We report the averaging performance on all tasks. We implement two versions of the \textsc{Full Attn.}, which differ in whether including left-to-right attention of the query.}
\label{tab:attn}
\vspace{-1mm}
\end{table*}

%% file: tables/window.tex
% \begin{figure}[ht]
%   \centering
%   \begin{subfigure}[c]{\linewidth}
%       \includegraphics[width=1.\linewidth]{figures/window.pdf}
%       \label{fig:res}
%   \end{subfigure}
%   \begin{subfigure}[c]{\linewidth}
%       \centering
%       \begin{tabular}{lcccc}
%         \toprule
%         Window & n-shots & LLAMA-65B & LLAMA-30B \\
%         \toprule
%         regular & 0-1-2-3-4 & 69.87 & 63.10 \\
%         \midrule
%         W=1 & 0-0-0-0-4 & 60.83 & 52.78 \\
%         W=2 & 1-1-1-1-4 & 71.33 & 60.83 \\
%         W=3 & 2-2-2-2-4 & 74.01 & 65.70 \\
%         W=4 & 3-3-3-3-4 & 74.27 & 68.73 \\
%         \bottomrule
%       \end{tabular}
%   \end{subfigure}
%   \caption{Visualization of attention masks for different window sizes where $K=4$.}
% \end{figure}

\begin{figure}[th]
  \centering
  \includegraphics[width=1.\linewidth]{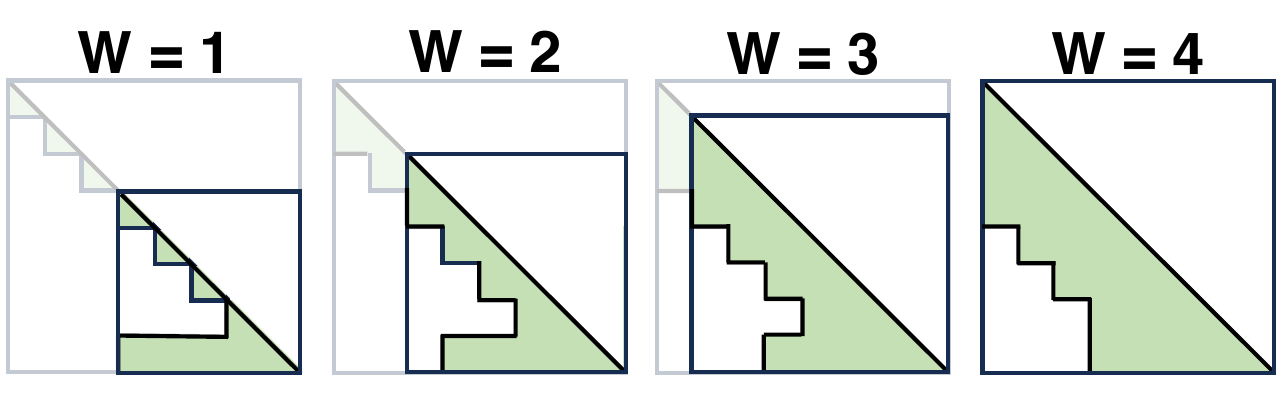}
  \caption{Visualization of attention masks for different window sizes.}
  \label{fig:window}
\end{figure}
\begin{table}[h]
    \centering
    \centering
      \scalebox{0.85}{\begin{tabular}{lcccc}
        \toprule
        Window & n-shots & LLAMA-65B & LLAMA-30B \\
        \toprule
        \textsc{Regular} & 0-1-2-3-4 & $69.87$ & $63.10$ \\
        \midrule
        W=1 & 0-0-0-0-4 & $60.83$ & $52.78$ \\
        W=2 & 1-1-1-1-4 & $71.33$ & $60.83$ \\
        W=3 & 2-2-2-2-4 & $74.01$ & $65.70$ \\
        W=4 & 3-3-3-3-4 & $74.27$ & $68.73$ \\
        \midrule
        W=4 & 3-3-3-3-3 & $70.99$ & $64.44$ \\
        \bottomrule
      \end{tabular}}
    \caption{Average performance on all tasks for different attention window sizes. We also report the number of demonstrations contained in the context of each demonstration. For example, "0-1-2-3-4" means that, for regular ICL, the first demonstration has no context, the second sees one demonstration, the third sees two, and so on. In this experiment, we use \textsc{KATE} to retrieve demonstrations instead of random-selected ones. In this way, we save computational resources because there is no need to run experiments over different random seeds.}
    \label{tab:window}
\end{table}

%% file: tables/appendix_full.tex
\begin{table}[h]
\begin{center}
\setlength{\tabcolsep}{1.2mm}{
\scalebox{0.8}{\begin{tabular}{lllll}
\toprule
Methods & 65B & 30B & 13B & 7B \\
\midrule
\textsc{Regular ICL} & 59.82 & 55.18 & 47.79 & 49.72 \\
\textsc{Rd} & 63.66 & 59.91 & 49.68 & 49.34 \\
\textsc{RdSca} & 75.76 & 69.32 & 53.03 & 50.92 \\
\midrule
\textit{with demonstration retrieval} \\
\textsc{Regular ICL} &  74.75 & 65.62 & 58.00 & 56.88 \\
\textsc{Rd} &  77.13 & 73.63 & 56.38 & 59.25 \\
\textsc{RdSca} &  87.38 & 82.13 & 68.88 & 64.63 \\
\bottomrule
\end{tabular}}}
\end{center}
\caption{Full evaluation results on SST-2}
\vspace{-1mm}
\end{table}

\begin{table}[h]
\begin{center}
\setlength{\tabcolsep}{1.2mm}{
\scalebox{0.8}{\begin{tabular}{lllll}
\toprule
Methods & 65B & 30B & 13B & 7B \\
\midrule
\textsc{Regular ICL} &  76.79 & 58.15 & 44.98 & 50.67 \\
\textsc{Rd} &  76.12 & 56.49 & 42.94 & 48.34 \\
\textsc{RdSca} &  76.90 & 56.17 & 46.97 & 50.66 \\
\midrule
\textit{with demonstration retrieval} \\
\textsc{Regular ICL} &  78.12 & 64.73 & 45.54 & 57.59 \\
\textsc{Rd} &  79.72 & 60.38 & 49.06 & 59.91 \\
\textsc{RdSca} &  79.72 & 60.38 & 52.83 & 57.08 \\
\bottomrule
\end{tabular}}}
\end{center}
\caption{Full evaluation results on CB}
\vspace{-1mm}
\end{table}

\begin{table}[h]
\begin{center}
\setlength{\tabcolsep}{1.2mm}{
\scalebox{0.8}{\begin{tabular}{lllll}
\toprule
Methods & 65B & 30B & 13B & 7B \\
\midrule
\textsc{Regular ICL} &  53.67 & 49.13 & 61.50 & 59.62 \\
\textsc{Rd} &  53.20 & 53.37 & 62.54 & 57.96 \\
\textsc{RdSca} &  59.68 & 53.44 & 60.05 & 57.23 \\
\midrule
\textit{with demonstration retrieval} \\
\textsc{Regular ICL} &  60.17 & 53.85 & 67.94 & 63.47 \\
\textsc{Rd} &  61.79 & 62.95 & 70.38 & 64.20 \\
\textsc{RdSca} &  64.87 & 61.54 & 66.79 & 64.36\\
\bottomrule
\end{tabular}}}
\end{center}
\caption{Full evaluation results on QQP}
\vspace{-1mm}
\end{table}

\begin{table}[h]
\begin{center}
\setlength{\tabcolsep}{1.2mm}{
\scalebox{0.8}{\begin{tabular}{lllll}
\toprule
Methods & 65B & 30B & 13B & 7B \\
\midrule
\textsc{Regular ICL} &  59.38 & 53.45 & 47.84 & 46.56 \\
\textsc{Rd} &  58.27 & 52.28 & 47.77 & 47.63 \\
\textsc{RdSca} &  60.51 & 54.49 & 45.33 & 47.29 \\
\midrule
\textit{with demonstration retrieval} \\
\textsc{Regular ICL} &  63.21 & 60.90 & 51.79 & 52.56 \\
\textsc{Rd} &  62.03 & 65.67 & 51.66 & 53.14 \\
\textsc{RdSca} &  64.43 & 66.15 & 51.27 & 52.98 \\
\bottomrule
\end{tabular}}}
\end{center}
\caption{Full evaluation results on QNLI}
\vspace{-1mm}
\end{table}

\begin{table}[h]
\begin{center}
\setlength{\tabcolsep}{1.2mm}{
\scalebox{0.8}{\begin{tabular}{lllll}
\toprule
Methods & 65B & 30B & 13B & 7B \\
\midrule
\textsc{Regular ICL} &  36.97 & 36.33 & 29.69 & 28.19 \\
\textsc{Rd} &  37.86 & 36.49 & 30.31 & 27.64 \\
\textsc{RdSca} &  43.85 & 40.09 & 31.12 & 30.89 \\
\midrule
\textit{with demonstration retrieval} \\
\textsc{Regular ICL} &  70.75 & 70.38 & 67.12 & 67.50 \\
\textsc{Rd} &  74.63 & 76.12 & 72.00 & 74.12 \\
\textsc{RdSca} &  75.13 & 76.25 & 70.12 & 70.75 \\
\bottomrule
\end{tabular}}}
\end{center}
\caption{Full evaluation results on AG-News}
\vspace{-1mm}
\end{table}

\begin{table}[h]
\begin{center}
\setlength{\tabcolsep}{1.2mm}{
\scalebox{0.8}{\begin{tabular}{lllll}
\toprule
Methods & 65B & 30B & 13B & 7B \\
\midrule
\textsc{Regular ICL} &  72.36 & 63.99 & 55.4 & 53.76 \\
\textsc{Rd} &  72.0 & 65.56 & 55.4 & 55.6 \\
\textsc{RdSca} &  73.07 & 65.8 & 57.07 & 53.95 \\
\midrule
\textit{with demonstration retrieval} \\
\textsc{Regular ICL} &  72.25 & 63.12 & 56.75 & 52.75 \\
\textsc{Rd} &  72.39 & 65.27 & 55.7 & 54.66 \\
\textsc{RdSca} &  74.07 & 65.92 & 56.74 & 53.23 \\
\bottomrule
\end{tabular}}}
\end{center}
\caption{Full evaluation results on RTE}
\vspace{-1mm}
\end{table}